# On the origins of self-modeling

Robert Kwiatkowski, Yuhang Hu, Boyuan Chen, Hod Lipson

## Abstract

Self-Modeling is the process by which an agent, such as an animal or machine, learns to create a predictive model of its own dynamics. Once captured, this self-model can then allow the agent to plan and evaluate various potential behaviors internally using the self-model, rather than using costly physical experimentation. Here, we quantify the benefits of such self-modeling against the complexity of the robot. We find a $R^2$=0.90 correlation between the number of degrees of freedom a robot has, and the added value of self-modeling as compared to a direct learning baseline. This result may help motivate self modeling in increasingly complex robotic systems, as well as shed light on the origins of self-modeling, and ultimately self-awareness, in animals and humans.

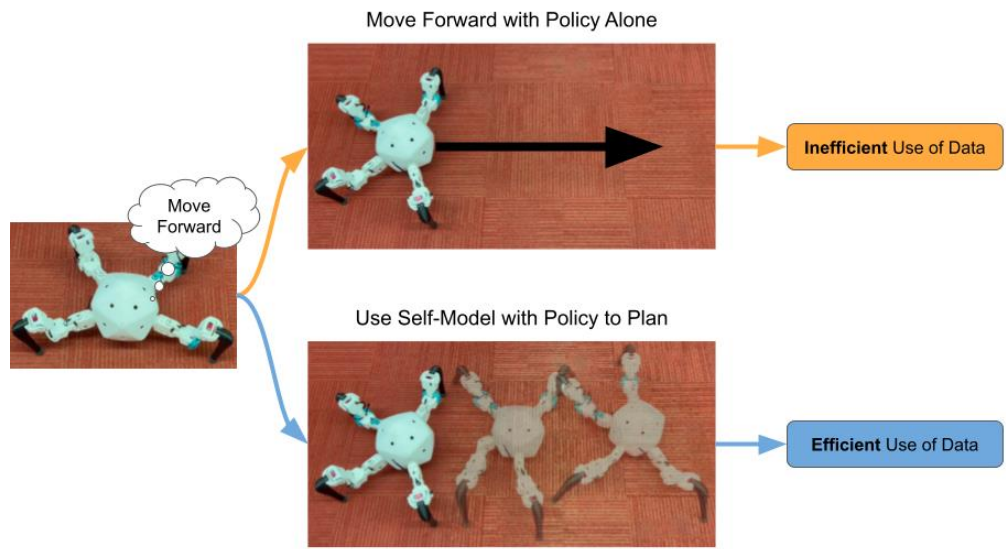

**Fig 1:** Overview of learning and deployment methodology of Dyna style algorithms

## Introduction

    Self-Modeling in humans, animals, and robots alike is the process whereby the agent learns to model its own dynamics. Simple self-models can be used to predict the immediate consequences of actions, such as motor commands. Such short-term mechanical self-models are known as forward-kinematics models. More complex self-models can be used to make

longer term predictions, such as predicting expected sensations and outcomes that are likely to result from potential actions, policies, and behaviors.

Self-modeling is long understood as essential for adaptive behavior both for machines and animals. Self-models have been used to allow robots to recover from unforeseen situations and recover from damage, such as loss of a leg[1–3]. Self modeling has also been shown to be a critical element in the effective planning of complex motions in animals[4,5]. These studies confirm that the use of a predictive forward model is crucial for long term planning in complex animals such as vertebrates. Consequently, we hypothesize that Self-Models, when used in a robotics context, would also become increasingly beneficial as the complexity of the robot being modeled increases.

To test this hypothesis, we use standard Reinforcement Learning (RL) implementation as a tool to test the utility of Self-Models. Reinforcement Learning has been successful in many challenging control problems[6–9]. In essence, Reinforcement Learning captures statistical relationships between current states, actions, and future states. These successes, however, are seldom realized in the real world because RL algorithms require a great deal of data to complete the statistical training, a problem known as Sampling Efficiency. Sampling efficiency a problem that Self-Models can help address, since once acquired, self-models provide an internal "surrogate" for predicting future states using the output from RL agents, thus replacing the need for extensive testing in reality. Our reasoning is that for the same amount of data, self-models capture information in a more useful and efficient way than model-free reinforcement learning.

In this work we find that the usefulness of these Self-Models increases linearly with the complexity of the robot. While there are many ways to quantify the complexity of a robot, we chose to quantify the complexity of the body being modeled as the number of mechanical degrees of freedom it has. We then did a side by side comparison where robots with various complexities used data to either train a reinforcement learning model, or train a self-model and then use the self model for training reinforcement learner. Other hybrid combinations could be explored in the future.

# Related Work

## Learning Robot Dynamics

Work has been done in the past on learning robot dynamics and leveraging those models to make robots accomplish their tasks. Bongard et al.[1] worked on a quadrupedal walking robot, however notably a quadruped with a less intuitive motor layout. This work created an algorithm for generating a physical model of the quadruped robot. The model they created was sufficient to generate a walking gait. However, this paper crucially included a simulator with which to generate the physical model. This simulator requires a knowledge of the physics and mechanics of the world and so is not a fully data driven approach. Kwiatkowski and Lipson[10] also worked to generate a self-model of a robot that was sufficient to do planning on. While the model was accurate enough to do a variety of tasks the robot used was far less complex and no task was learned. Amos et al.[11] too used machine learning to model the dynamics of a robotic system. For this paper the authors chose a robotic hand, a platform with significant complexity. However, this paper did not test their models as rigorously as the other works as they did not use their model for a variety of complex tasks.

## Limited Action Space Environment Learning

Learning environments in step with reinforcement learning is popular due to its great successes. Ha et al.[12] successfully leverage the use of models of reinforcement learning environments to train agents. This work however fully integrated their world models with the agents so that the agent would make use of the latent space of the world model as opposed to some predicted state $\hat{S}$. Chiappa et al. [13] also used recurrent models to learn a model of their environment. This work does an extensive review of a number of different ways to effectively predict the environment as well as work on how to integrate it with reinforcement learning in the discrete action space. The authors touch on some elements that translate well to using environment learning to create self-models such as the learning of prediction independent simulators. Racanière et al.[14] also proposes an architecture for leveraging environment learning to make a more efficient reinforcement learning algorithm. This paper showed success in the Sokoban environment however it still had a limited dimensional action space and as such much simpler world dynamics. Pan et al.[15] similarly explored methods for merging model based and model free reinforcement learning together to solve discrete action space reinforcement learning problems. Kaiser et al.[16] gathered data about the world and then used a world model to train a reinforcement learner.

# Experimental Design

For all of the experiments in this paper we used a simulated robot to collect data through random motion. This data was later used to train a self-model on which the task planner learned to execute a task without any additional observations from the simulator. All experiments were conducted on an Nvidia 1080 TI GPU.

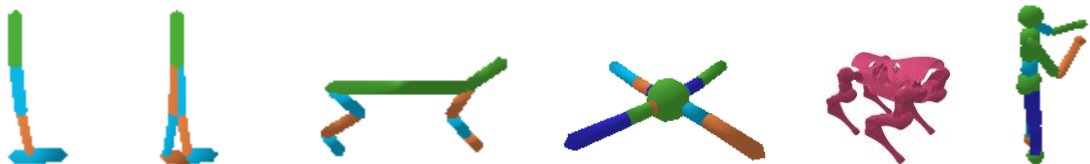
Figure 3: The 6 robots used to demonstrate the power of Self-Models.

## Pybullet Environments

For this work, we collected all data and ran all experiments on the PyBullet[17] simulation's varied robot morphologies as seen in Figure 3. These robots have been used in simulation in the past and have proven themselves to be an interesting platform to work with. For example Ant has shown itself to be sufficiently versatile that it has been able to run[18], as well as jump and navigate obstacles[19], while the Humanoid has shown itself to be a relatively difficult control task.

For all of these robot morphologies, the actions exist in the range [-5, 5] and so in our experiments we normalize them to fall within [-1, 1]. The state space however is more complex. Each state contain *2m+f+8* continuous values, where *m* corresponds to the number of motors and *f* corresponds to the number of feet, This state space corresponds to the change in *z* position, the sine and cosine of the angle of the robot to the predefined "target" position, the velocities in the x, y, and z direction, the roll and pitch, the speed and position of each of the joints, as well as information regarding the feet contacts.

For the purposes of our experiments however, we were able to ignore all of the task specific information regarding the target position and focus on the sensory input alone. This allows us to be as general as possible including only the speed and position of each of the simulated motors as well as the *x,y,z* velocity measurements and the roll and pitch measurements produced by the simulator as our state space, all sensor measurements that would be very commonly found on real robots.

## Task Planner

In order to transfer the general knowledge of the self contained in the self-model to a more specific knowledge of tasks we use Proximal Policy Optimization (PPO)[18,19] to train a policy for each task. PPO has been used in the past to learn robotic control policies in simulation and in the real world alike. We use PPO to learn a policy for controlling the robot by just using the self-model trained on the robot's movements. Our PPO agent has a policy network consisting of 2 fully connected layers of 64 hidden units each. This network while not particularly complex was sufficient to learn all our intended policies on real data and as such is sufficient for all our experiments.

In the normal reinforcement learning paradigm, a reinforcement learning agent is given a state and learns an appropriate action in order to maximize its reward over an episode. Normally, this state comes from the "real world," sometimes data from a real robot, but more often data from a hand-coded simulator. In this work, we instead give the data output from the self-model.

In this paper we ran 2 experiments, both of which had the exact same setup except for the calculation of the reward function. In every experiment the first thing that is done is the simulated robot is "reset" to an initial state and that first observation is returned. This observation $S_0$ is given to the self-model as a seed observation and it is given to the agent as well. The agent then produces its first action $A_0$. This action is passed to the self-model who will then output the predicted next state $\hat{S}_1$. $\hat{S}_1$ is then passed to the agent who will produce $A_1$. This process will continue until the episode length is reached.

For the 2 different experiments we changed the reward function in order to cause the agent to learn a new task. Unlike the state, we instead used the true reward function as the learning of reward functions is outside the scope of this research. The first task we learned was forward locomotion. In order to learn this task we use the *x* velocity at each timestep as the reward so that the episode's final reward is the total distance traveled resulting in agents who move forward efficiently. The second task we learned was jumping, or vertical locomotion. For this task, we set our reward equal to the velocity in the *z* direction with the added effect that if the agent passed a defined *z* position then the episode would terminate. This was done for smoothness in the agent's learning.

## Self-Model Utility Scales with Complexity

To quantify the relationship between robot complexity and usefulness of self-models we compare the performance of the baseline MFRL agent trained using a dataset *D* to that of the Self-Model based approach where the Self-Model was trained also using |*D*| states. We then measured how well the Self-Model based approach performed as a percentage of the baseline and repeated this process for each of the different DoF robots; these results can be seen in Figure 2. We use DoF as a proxy for complexity of the robot which correlates strongly with complexity.

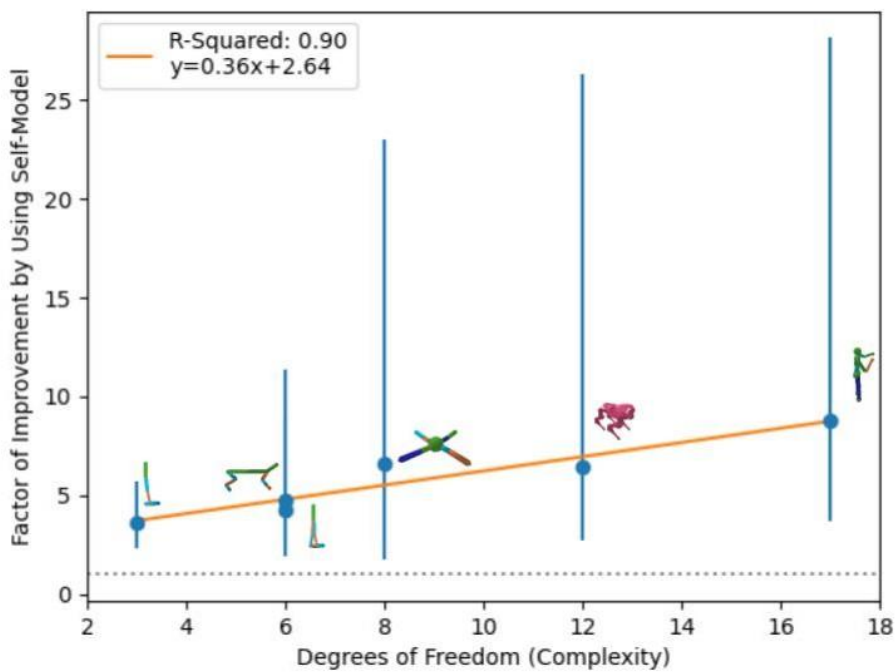

**Fig 2:** Correlation between task complexity and the utility of Self-Models at n=10 samples. We also depict the 6 different robot types of varying complexity used in our experiments.

Once we have collected measurements for each body type, we calculate the Coefficient of determination (R-Squared) between the DoF and percent improvement from using the Self-Model. This coefficient of determination measures how strongly correlated are the improvement in using a Self-Model, and thus the usefulness of the Self-Model, and the complexity of a task being executed. We find that at |*D*|=1000 there is an R-Squared=0.9 suggesting that 90% of this improvement can be explained purely as a result of the increase in complexity. This exceptionally strong correlation shows that as the complexity of problems increases the more value is derived from using a Self-Model, especially at the low data cases.

To validate our results outside of simulation we also tested the MFRL and Self-Model based approaches on a real robot of 12 DoF. For this experiment we used |D|=6000 and trained the self-model on data collected using the robot's onboard sensors as well as a motion capture

system to precisely capture the position and orientation of the robot. We reset the robot after each episode was completed and ran the robot to 100 steps each episode. Once trained we repeated the same process as outlined above but using the data from the real robot to train the Self-Model.

Once the policy was trained we deployed it to the real robot again and observed the results. These results can be seen in our supplementary material S1 and S2 and shows that the experiments using self-models far outperform the experiments where they are not used. This experiment showed an improvement of 3000% over that of an RL agent trained with the same data which is within the expected error bounds predicted by our simulated experiments. This success suggests that the observed phenomenon is not just limited to experimental experiments and that when deployed on real systems of increasing complexity so too does the value of the Self-Model.

This result was further supported by the fact that these Self-Model based experiments were then replicated by using a more advanced Dyna style algorithm (MBPO[20]). When using MBPO with a self-model we observed improvements far in excess of the predicted improvements in Figure 1. This is to be expected as MBPO is a much more complicated and efficient algorithm than the baseline Dyna approach used in all prior experiments. Such a result gives credence to the hypothesis that our observed result will hold even through the use of more complex MBRL algorithms.

# Results

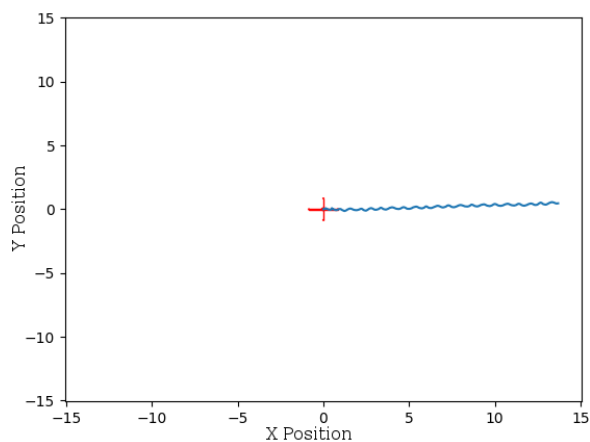
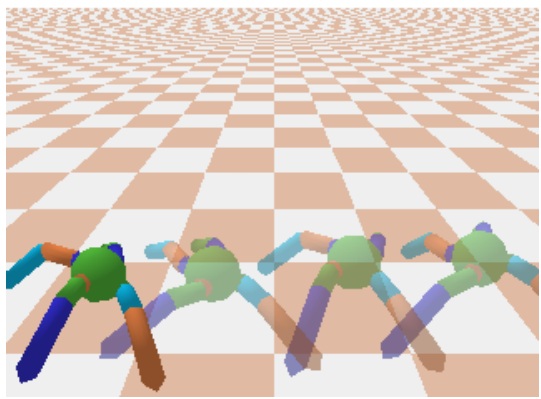

a) Gait Trajectory         b) Gait Visualization

Figure 4: Walking policy as learned by the self-model. Figure 4(a) shows the distance traveled (blue) as compared to the body of the Ant (red). Figure 4(b) shows the same policy.

## Learning to Walk

To gauge the performance of Self-Modeling at higher data levels, we also tested our algorithm using the same architecture and model configuration on a variety of different robot platforms found in the PyBullet API and demonstrated a significant improvement in data efficiency on every platform tested. For every experiment we trained the self-model with up to 100,000 real state action tuples whereas we trained the PPO agents to up to 1,000,000 state action tuples. When using comparable data the policy trained on self-model artificial data outperformed the PPO agent trained on real data. This further suggests that the ability of the self-model to generalize on small amounts of data is very strong.

Despite training entirely in self-model, the agents learn an effective gait and one that can translate well into the real world. Figure 4 shows that the path traveled by the Ant robot, shown in blue, far exceeds the size of the robot shown as the red cross. Similarly, the path here is almost entirely straight with only minor deviations in a wave-like pattern. These deviations are to be expected as the position reported corresponds to the center of the robot, and while moving it has a tendency to sway back and forth.

This gait itself is very similar to those trained using the real simulator. The learned gait also progressed in a fairly natural way, and one similar to those gaits learned on real simulators. At 100,000 steps the policy had begun to learn how to catch itself from falling in an effective manner, and that moving one of its legs is a useful strategy. It first moves its one leg back and forth and can be seen in dark blue. At 500,000 steps the policy has also noticeably improved. The robot has learned to make use of all 4 of its legs to land and immediately start to take its first steps. However, after catching itself from falling the robot resumes the inefficient policy of moving one leg back and forth. As the agent approaches 1,000,000 steps the gait becomes more general. Instead of taking one step in the beginning of the episode the agent has learned how to take a step and from that position take another generalizing into a complete walking gait. This timeframe to achieve success mirrors the process that the agent learning in the real environment takes, suggesting that the self-model is a close approximation to the real environment.

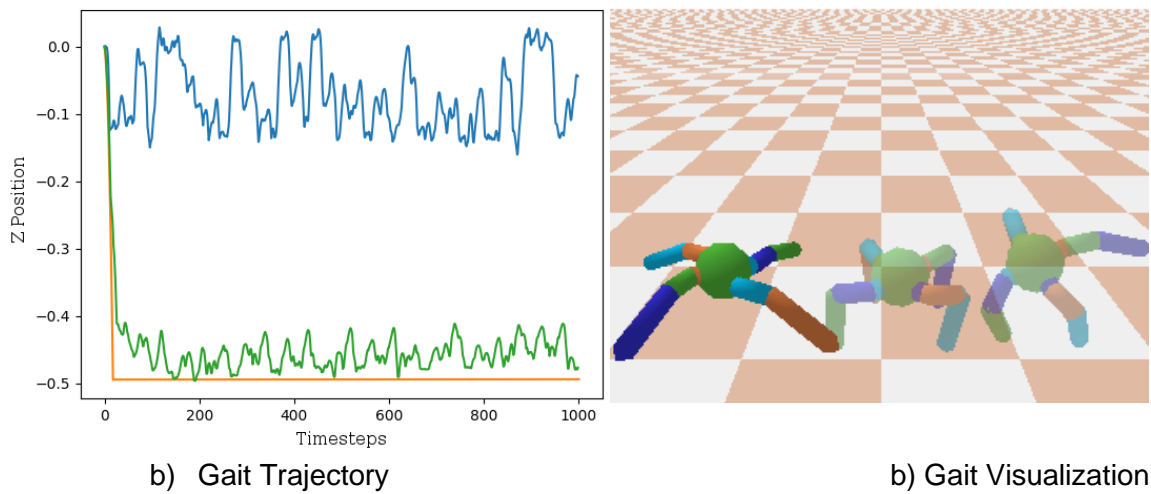

b) Gait Trajectory        b) Gait Visualization

Figure 5: Jumping policy as learned by the self-model. Figure 5(a) shows the distance traveled (blue) as compared to the same policy applied without a self-model (green) and with no policy (orange). Figure 5(b) shows the motion of the same policy.

## Learning to Jump

One of the most important benefits of the self-model based approach to training agents is its ability to immediately shift to learning a new task without the need to collect any new state action tuples. Our trained agent showed that it was able to learn a jumping policy using the self-model in an open loop. The agent learns a policy that is noticeably different from the walking policies. The agent first catches itself with its legs much straighter than before and then lifts the legs off of the ground to produce some momentum such that it can eventually propel itself upward and then fall gracefully back down to the ground so that it can do it again.

The success of the jumping policy is even more notable when the motion in the vertical ($z$) direction is plotted over the duration of the episode. In Figure 5 we show the $z$ motion for the jumping policy trained on the self-model, the running policy trained on the self-model and an untrained policy as a baseline. The blue chart (the policy trained to jump) shows not only a significantly higher average $z$ position but a significantly higher variance than both of the other policies. The higher average $z$ position shows it has successfully achieved its goal outlined by its reward function which rewarded it for moving higher. The maximum heights shown also suggests that the agent has successfully learned to jump. The agent in this situation would not be able to go higher than a height of 0 with its feet on the ground. The successes shown in this experiment further outline the power of the self-model which was able to go seamlessly from learning to walk to learning a completely new and separate task.

## Conclusion

Both self-modeling and conventional Reinforcement Learning use data to train a predictive dynamical model, but they use the data in substantially different ways. Whereas RL uses data to model the self and the task in a combined statistical model, self-modeling separates the "self" in a way that is more task-agnostic, reusable and efficient. We suggest that simpler organisms that perform a relatively small and finite set of tasks in a limited set of environments, can "get away" with one large behavioral model that is relatively unadaptive. In contrast, more complex organisms with many degrees of freedom and larger range of tasks, cannot compress their knowledge into a single model. Instead, they benefit from separating a statistical model of self from a model of everything else. This separation confers a learning advantage that grows with the complexity of the system. Finally, we suggest that this statistical learning advantage translated into an evolutionary advantage that may have contributed to the origin of self-awareness in humans.

# List of Supplementary Material

S1: Demonstration of the difference between an RL agent trained with and without the use of a trained Self-Model
S2: Demonstration of the difference between a classical control agent with and without the use of a trained Self-Model
S3: Diagram of the robots used in testing
S4: Detailed sketch of the robots used in testing

# Supplementary Material:

## S1

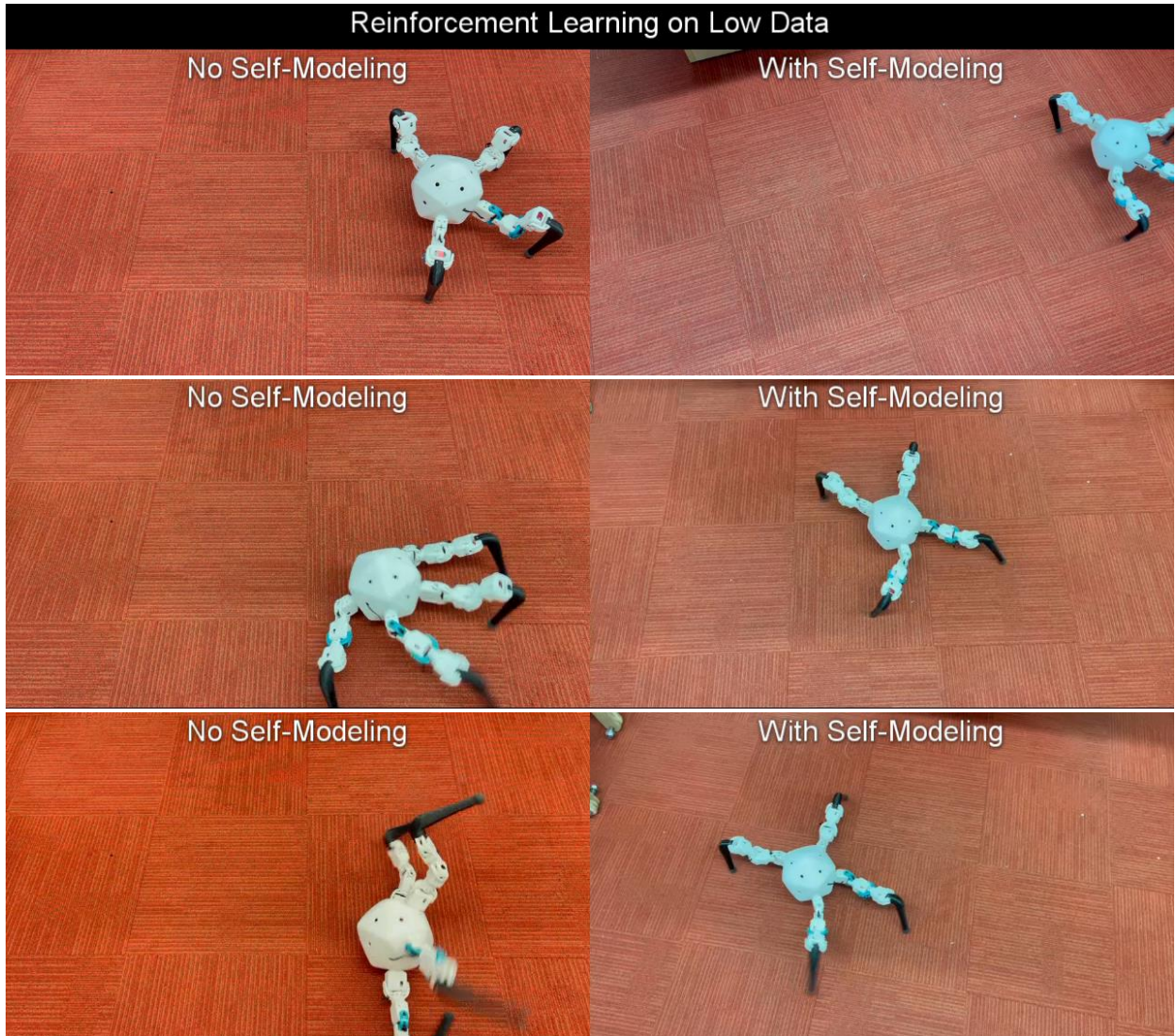

    This video demonstrates the significant difference between the usage of reinforcement learning through the Dyna approach with and without Self-Modeling on a 12 DoF quadrapedal robot. In the attached video it can be seen that the robot with Self-Modeling is able to walk forward successfully while the robot without self-modeling is unable to accomplish the task and instead struggles and never actually succeeds to walk forward.

# S2

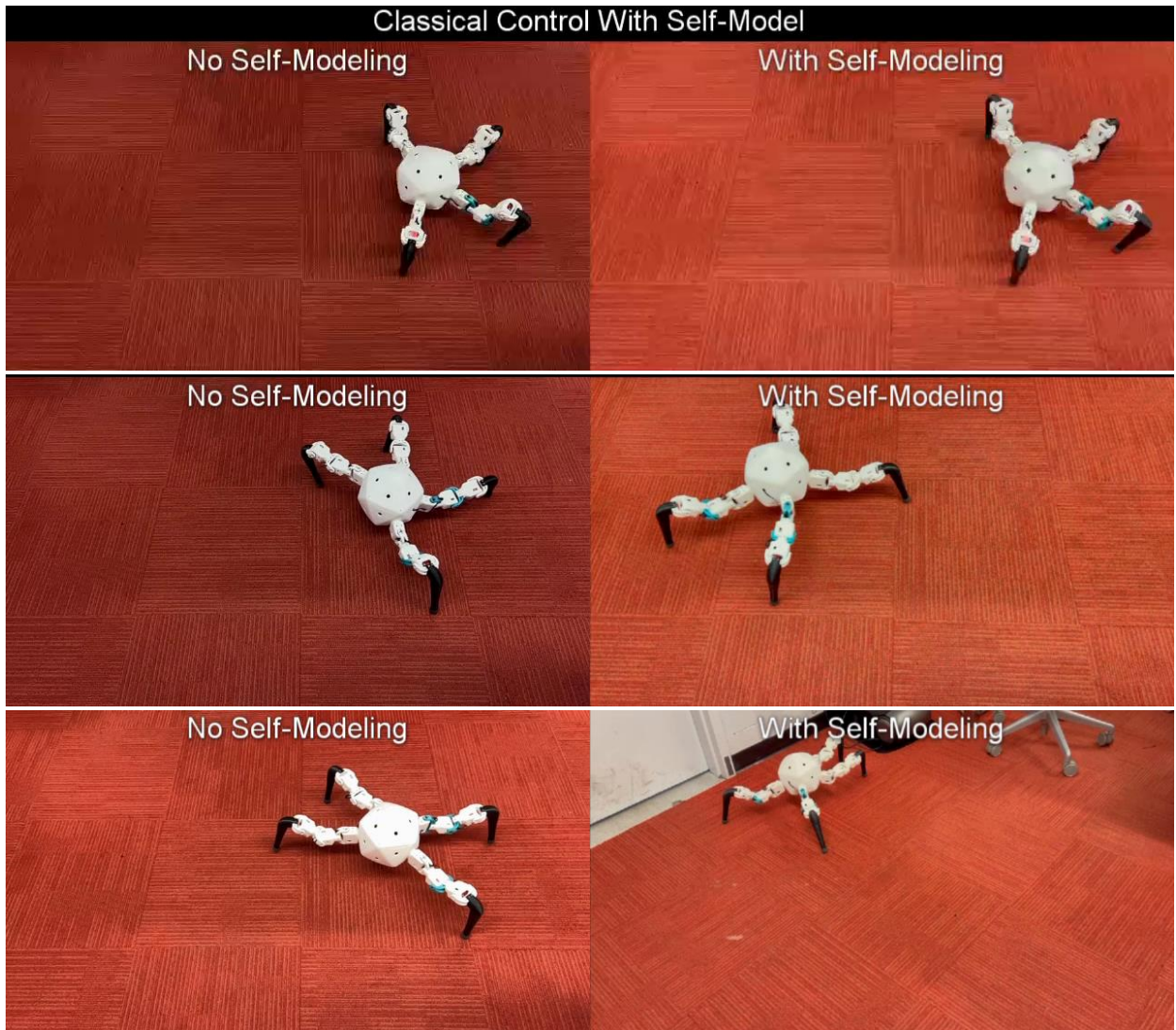

      This video demonstrates the difference between the use of classical control algorithms such as a sinusoidal gait generator with and without a Self-Model. Although the differences are not as pronounced as in the case of reinforcement learning. It is apparent that the Self-Modeling gait performs notably better than the gait without the use of a Self-Model. The Self-Modeling robot is able to walk forward successfully with minimal deviation from the course whereas the robot without Self-Modeling turns significantly more and makes much less forward progress.

# S3

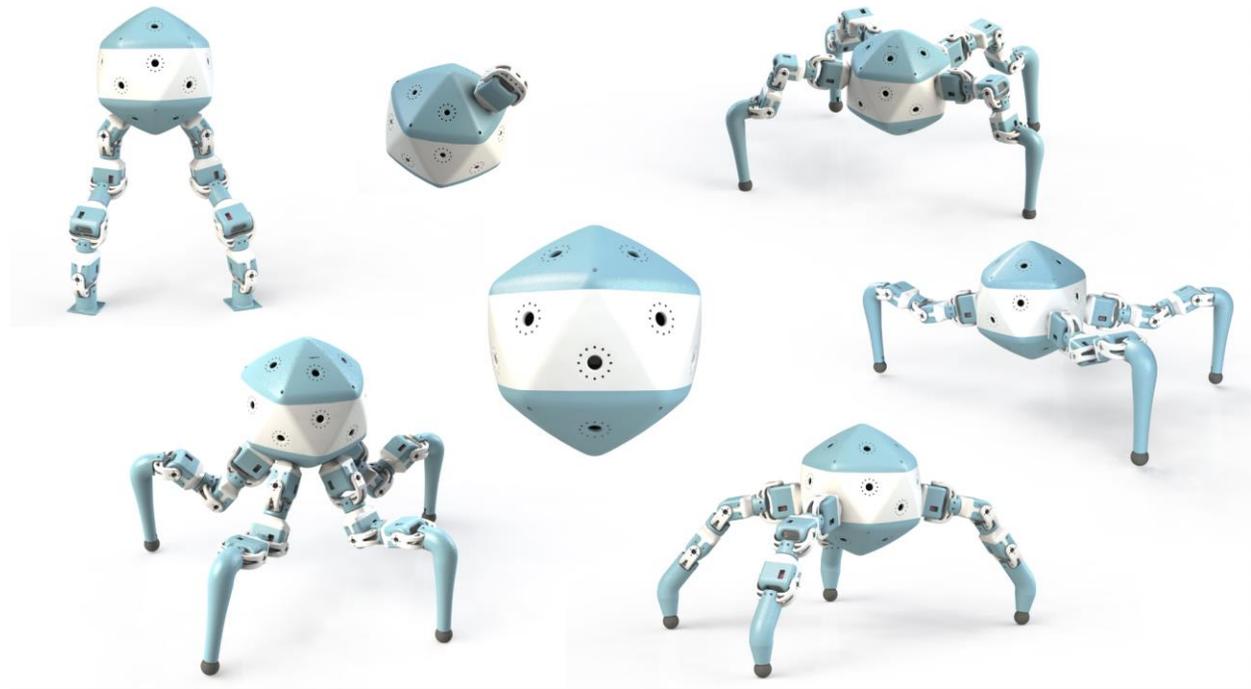

    This image contains examples of the robot used for physical testing. The robot was placed in a quadrapedal form and has 3 motors on each of its 4 legs. This allows the robot to be tested in a 12 DoF as well as an 8 DoF and 4 DoF configuration if needed.

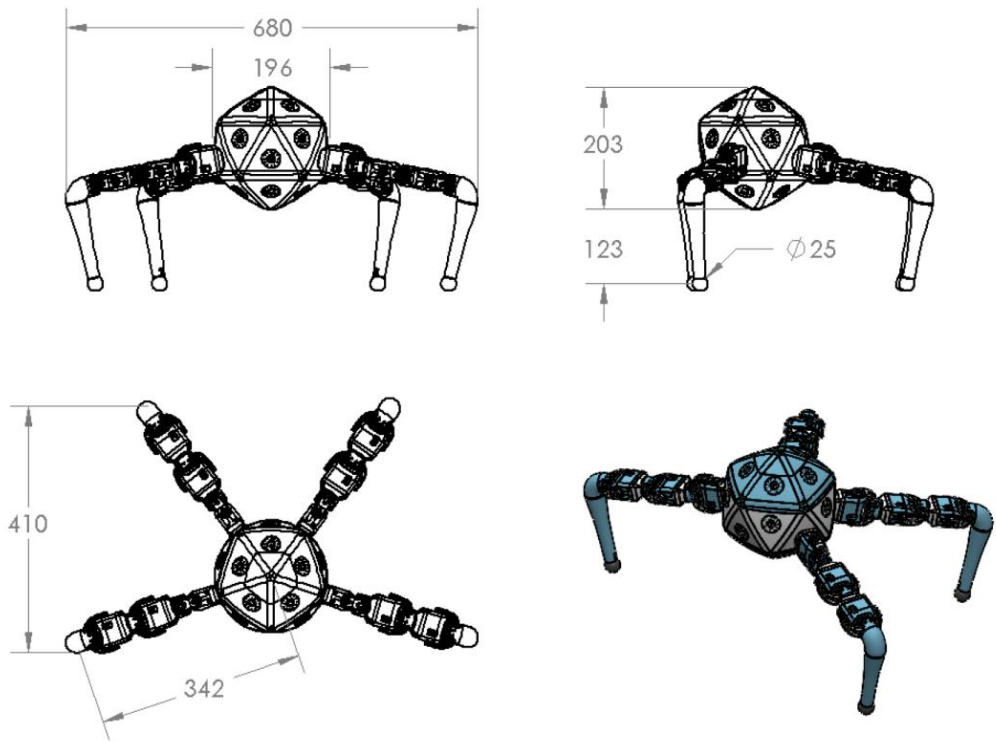

A detailed sketch of the dimensions of the robot used in testing. This image shows all relevant distances necessary to reproduce the robot built.